\documentclass{article} 
\usepackage{iclr2017_conference,times}
\usepackage{hyperref}
\usepackage{url}
\usepackage{color}
\usepackage{layouts}
\usepackage{float}
\usepackage{graphicx}
\usepackage{amsmath}
\usepackage{amsfonts}
\usepackage{color}
\usepackage{inputenc}
\usepackage{mathtools}
\usepackage{algpseudocode}
\usepackage{algorithm}
\usepackage{multirow}
\usepackage{array}

\newcolumntype{L}[1]{>{\raggedright\let\newline\\\arraybackslash\hspace{0pt}}m{#1}}
\newcolumntype{C}[1]{>{\centering\let\newline\\\arraybackslash\hspace{0pt}}m{#1}}
\newcolumntype{R}[1]{>{\raggedleft\let\newline\\\arraybackslash\hspace{0pt}}m{#1}}

\usepackage{caption} 
\title{Neural Combinatorial Optimization \\with Reinforcement Learning}

\author{Irwan Bello\thanks{Equal contributions. Members of the Google Brain Residency program (\url{g.co/brainresidency}).}~, Hieu Pham\footnotemark[1]~, Quoc V. Le, Mohammad Norouzi, Samy Bengio \\
Google Brain\\
\texttt{\{ibello,hyhieu,qvl,mnorouzi,bengio\}@google.com}}

\usepackage{amsthm}
\usepackage{amsmath}
\usepackage{amssymb}
\usepackage{color}
\usepackage{xspace}
\usepackage{pgfplots, pgfplotstable}

\newcommand{\comment}[1]{}

\def\prerl{RL pretraining\xspace}
\def\as{active search\xspace}

\renewcommand{\vec}[1]{\boldsymbol{\mathbf{#1}}}

\def\btheta{\vec{\theta}}

\def\bx{\vec{x}}

\def\int{\mathrm{int}}

\def\ie{{\em i.e.}}

\newcommand{\tabref}[1]{Table~\ref{#1}}
\newcommand{\figref}[1]{Figure~\ref{#1}}

\newcommand{\vectornorm}[1]{\left\| #1 \right\|}
\newcommand{\defeq}{\stackrel{\mathclap{\normalfont\mbox{\tiny def}}}{=}}

\begin{document}

\maketitle
\begin{abstract}

This paper presents a framework to tackle combinatorial optimization
problems using neural networks and reinforcement learning. We focus on
the traveling salesman problem (TSP) and train a recurrent neural network
that, given a set of city \mbox{coordinates}, predicts a distribution
over different city permutations. Using negative tour length as
the reward signal, we optimize the parameters of the recurrent neural network
using a policy gradient method. We compare learning the network
parameters on a set of training graphs against learning them on
individual test graphs. Despite the computational expense, without much
engineering and heuristic designing, Neural Combinatorial Optimization achieves 
close to optimal results on 2D Euclidean graphs with up to $100$ nodes. 
Applied to the KnapSack, another NP-hard problem, the same method obtains 
optimal solutions for instances with up to $200$ items.

\end{abstract}

\section{\label{sec:intro}Introduction}

{\em Combinatorial optimization} is a fundamental problem in computer science.
A canonical example is the {\em traveling salesman problem (TSP)}, 
where given a graph, one needs to search the space of permutations to find 
an optimal sequence of nodes with minimal total edge weights (tour length). 
The TSP and its variants have myriad applications in planning, manufacturing, 
genetics, {\em etc.}~(see \citep{tspbook} for an overview). 

Finding the optimal TSP solution is NP-hard, even in the two-dimensional
Euclidean case~\citep{papadimitriou77}, where the nodes are 2D points and edge
weights are Euclidean distances between pairs of points. 
In practice, TSP solvers rely on handcrafted heuristics that guide 
their search procedures to find competitive tours efficiently. Even though these 
heuristics work well on TSP, once the problem statement changes slightly, 
they need to be revised.
In contrast, machine learning methods have the potential to be 
applicable across many optimization tasks by automatically discovering their 
own heuristics based on the training data, thus requiring less hand-engineering
than solvers that are optimized for one task only.

While most successful machine learning techniques fall into the family of
supervised learning, where a mapping from training inputs to outputs is
learned, supervised learning is not applicable to most combinatorial
optimization problems because one does not have access to optimal labels.
However, one can compare the quality of a set of solutions using a verifier,
and provide some reward feedbacks to a learning algorithm. Hence, we follow
the reinforcement learning (RL) paradigm to tackle combinatorial optimization.
We empirically demonstrate that, even when using optimal solutions as labeled
data to optimize a supervised mapping, the generalization is rather poor
compared to an RL agent that explores different tours and observes their
corresponding rewards.

We propose Neural Combinatorial Optimization, a framework to tackle
combinatorial optimization problems using reinforcement learning and neural
networks. We consider two approaches based on policy gradients~\citep{reinforce}. 
The first approach, called {\em \prerl}, uses a training set to optimize a 
recurrent neural network (RNN) that parameterizes a stochastic policy over solutions, 
using the expected reward as objective.
At test time, the policy is fixed, and one performs inference by greedy
decoding or sampling. The second approach, called {\em \as}, involves no
pretraining. It starts from a random policy and iteratively optimizes the RNN
parameters on a single test instance, again using the expected reward
objective, while keeping track of the best solution sampled during the search.
We find that combining \prerl and \as works best in practice.

On 2D Euclidean graphs with up to $100$ nodes, Neural Combinatorial Optimization 
significantly outperforms the supervised learning approach to the 
TSP~\citep{vinyals2015pointer} and obtains close to optimal results when allowed 
more computation time. We illustrate its flexibility by testing the same method 
on the KnapSack problem, for which we get optimal results for instances with up to 200 items.
These results give insights into how neural networks can be used as a general tool 
for tackling combinatorial optimization problems, especially those that are difficult
to design heuristics for.

\section{\label{sec:previous}Previous Work}
 
The Traveling Salesman Problem is a well studied combinatorial optimization
problem and many exact or approximate algorithms 
have been proposed for both Euclidean and non-Euclidean graphs. 
\cite{christofides} proposes a heuristic algorithm that involves computing a 
minimum-spanning tree and a minimum-weight perfect matching. 
The algorithm has polynomial running time and returns solutions that are 
guaranteed to be within a factor of $1.5\times$ to optimality in the metric 
instance of the TSP. 

The best known exact dynamic programming algorithm for TSP has a complexity of
$\Theta(2^n n^2)$, making it infeasible to scale up to large instances, say
with $40$ points. 
Nevertheless, state of the art TSP solvers, thanks to
carefully handcrafted heuristics that describe how to navigate the space of
feasible solutions in an efficient manner, can solve symmetric TSP instances 
with thousands of nodes. Concorde \citep{concorde}, 
widely accepted as one of the best exact TSP solvers, makes
use of cutting plane algorithms \citep{dantzig1954solution, cutting,
applegate2003implementing}, iteratively solving linear programming relaxations
of the TSP, in conjunction with a branch-and-bound approach that prunes parts
of the search space that provably will not contain an optimal solution.
Similarly, the Lin-Kernighan-Helsgaun heuristic \citep{lkh}, 
inspired from the Lin-Kernighan heuristic \citep{lk},
is a state of the art approximate search heuristic for the symmetric TSP and 
has been shown to solve instances with hundreds of nodes to optimality.

More generic solvers, such as Google's vehicle routing problem 
solver \citep{OR-Tools} that tackles a superset of the TSP, 
typically rely on a combination of local search algorithms and metaheuristics.
Local search algorithms apply a specified set of local move operators 
on candidate solutions, based on hand-engineered heuristics such as 2-opt 
\citep{twoopt}, to navigate from solution to solution in the search space. 
A metaheuristic is then applied to propose uphill moves and escape local optima.
A popular choice of metaheuristic for the TSP and its variants is guided local
search \citep{guidedlocalsearch}, which moves out of a local minimum 
by penalizing particular solution features that it considers should not occur 
in a good solution.

The difficulty in applying existing search heuristics to newly encountered problems 
- or even new instances of a similar problem - is a well-known challenge that 
stems from the {\em No Free Lunch theorem}~\citep{nfl}.
Because all search algorithms have the same performance when averaged over all problems, 
one must appropriately rely on a prior over problems when selecting a search algorithm 
to guarantee performance.
This challenge has fostered interest in raising the level of generality at which 
optimization systems operate~\citep{burke2003} and is the underlying motivation 
behind hyper-heuristics, defined as "search method[s] or learning mechanism[s] 
for selecting or generating heuristics to solve computation search problems".
Hyper-heuristics aim to be easier to use than problem specific methods 
by partially abstracting away the knowledge intensive process of selecting 
heuristics given a combinatorial problem and have been shown to successfully 
combine human-defined heuristics in superior ways across many tasks 
(see~\citep{burke_survey} for a survey).
However, hyper-heuristics operate on the search space of heuristics, rather than 
the search space of solutions, therefore still initially relying on human created
heuristics.

The application of neural networks to combinatorial optimization has a
distinguished history, where the majority of research focuses on the Traveling
Salesman Problem~\citep{smith1999neural}. One of the earliest proposals is the
use of Hopfield networks~\citep{hopfield1985neural} for the TSP. 
The authors modify the network's energy function to make it equivalent to TSP 
objective and use Lagrange multipliers to penalize the violations of the problem's
constraints. A limitation of this approach is that it is sensitive to hyperparameters 
and parameter initialization as analyzed by~\citep{wilson1988stability}. 
Overcoming this limitation is central to the subsequent work in the field, especially
by~\citep{aiyer1990theoretical,gee1993problem}. 
Parallel to the development of Hopfield networks is the work on using deformable 
template models to solve TSP. Perhaps most prominent is the invention of Elastic Nets
as a means to solve TSP~\citep{durbin1987analogue}, and the application of 
Self Organizing Map to TSP~\citep{fort1988solving,angeniol1988self,kohonen1990self}. 
Addressing the limitations of deformable template models is central to the following 
work in this area~\citep{burke1994neural,favata1991study,vakhutinsky1995hierarchical}. 
Even though these neural networks have many appealing properties, 
they are still limited as research work. When being carefully benchmarked, 
they have not yielded satisfying results compared to algorithmic 
methods~\citep{sarwar2012critical,la2012comparison}. 
Perhaps due to the negative results, this research direction 
is largely overlooked since the turn of the century.

Motivated by the recent advancements in sequence-to-sequence
learning~\citep{sutskever2014sequence}, neural networks are again the subject 
of study for optimization in various domains~\citep{rnn_blackbox}, 
including discrete ones~\citep{barret}. In particular, the TSP is revisited 
in the introduction of Pointer Networks~\citep{vinyals2015pointer}, 
where a recurrent network with non-parametric softmaxes is 
trained in a supervised manner to predict the sequence of visited cities. 
Despite architecural improvements, their models were trained using 
supervised signals given by an approximate solver.
\section{\label{sec:methods}Neural Network Architecture for TSP}

We focus on the 2D Euclidean TSP in this paper. Given an input graph,
represented as a sequence of $n$ cities in a two dimensional space $s =
\{\bx_i\}_{i=1}^n$ where each $\bx_i \in \mathbb{R}^2$, we are concerned with
finding a permutation of the points $\pi$, termed a tour, that visits each city
once and has the minimum total length. We define the length of a tour defined
by a permutation $\pi$ as
\begin{align}
  L(\pi \mid s) = \vectornorm{\mathbf{x}_{\pi(n)} - \mathbf{x}_{\pi(1)}}_2
    + \sum_{i=1}^{n-1} \vectornorm{\mathbf{x}_{\pi(i)} - \mathbf{x}_{\pi(i+1)}}_2,
\end{align}
where $\vectornorm{\cdot}_2$ denotes $\ell_2$ norm.

We aim to learn the parameters of a stochastic policy $p(\pi \mid s)$ that
given an input set of points $s$, assigns high probabilities to short tours and
low probabilities to long tours. Our neural network architecture uses the chain
rule to factorize the probability of a tour as
\begin{align}
  p(\pi \mid s) = \prod_{i=1}^{n} p\left(\pi(i) \mid \pi(<i),
  s\right)~,
  \label{eqn:prob}
\end{align}
and then uses individual softmax modules to represent each term on the
RHS of \eqref{eqn:prob}.

We are inspired by previous work~\citep{sutskever2014sequence} that makes use
of the same factorization based on the chain rule to address sequence to
sequence problems like machine translation. One can use a vanilla sequence to
sequence model to address the TSP where the output vocabulary is $\{1, 2,
\ldots, n\}$. However, there are two major issues with this approach: (1)
networks trained in this fashion cannot generalize to inputs with more than $n$
cities. (2) one needs to have access to ground-truth output permutations to
optimize the parameters with conditional log-likelihood. We address both
isssues in this paper.

\begin{figure}[t]
    \centering
    \includegraphics[width=0.7\textwidth]{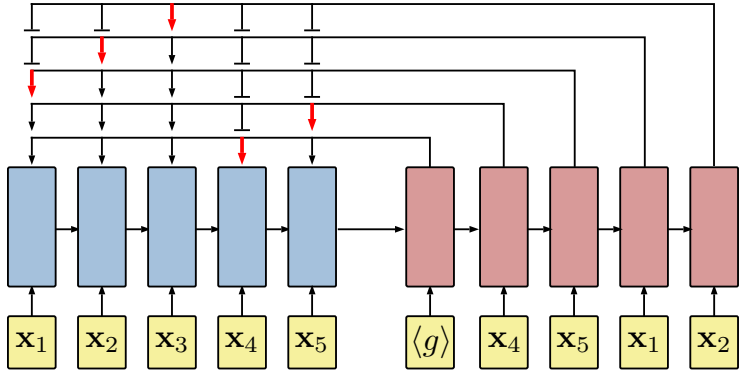}
    \caption{A pointer network architecture introduced by~\citep{vinyals2015pointer}. }
    \label{fig:ptrnet}
\end{figure}

For generalization beyond a pre-specified graph size, we follow the approach
of~\citep{vinyals2015pointer}, which makes use of a set of non-parameteric
softmax modules, resembling the attention mechanism from~\citep{bahdanau15}.
This approach, named {\em pointer network}, allows the model to effectively
point to a specific position in the input sequence rather than predicting an
index value from a fixed-size vocabulary. We employ the pointer network
architecture, depicted in~\figref{fig:ptrnet}, as our policy model to
parameterize $p(\pi \mid s)$.

\subsection{Architecture Details}
Our pointer network comprises two recurrent neural network (RNN) modules,
encoder and decoder, both of which consist of Long Short-Term Memory (LSTM)
cells~\citep{lstm97}. The encoder network reads the input sequence $s$, one
city at a time, and transforms it into a sequence of latent memory states
$\{enc_i\}_{i=1}^n$ where $enc_i \in \mathbb{R}^d$.  The input to the encoder
network at time step $i$ is a $d$-dimensional embedding of a 2D point $\bx_i$,
which is obtained via a linear transformation of $\bx_i$ shared across all
input steps.  The decoder network also maintains its latent memory states
$\{dec_i\}_{i=1}^n$ where $dec_i \in \mathbb{R}^d$ and, at each step $i$, uses
a pointing mechanism to produce a distribution over the next city to visit in
the tour. Once the next city is selected, it is passed as the input to the next
decoder step. The input of the first decoder step (denoted by $\langle g
\rangle$ in \figref{fig:ptrnet}) is a d-dimensional vector treated as a
trainable parameter of our neural network.

Our attention function, formally defined in Appendix~\ref{app:pointing}, takes
as input a query vector $q = dec_i \in \mathbb{R}^d$ and a set of reference
vectors $ref = \{enc_1, \ldots, enc_k\}$ where $enc_i \in \mathbb{R}^d$, and
predicts a distribution $A(ref,q)$ over the set of $k$ references.  This
probability distribution represents the degree to which the model is pointing 
to reference $r_i$ upon seeing query $q$.

\cite{vinyals2015order} also suggest including some additional computation
steps, named \textit{glimpses}, to aggregate the contributions of different
parts of the input sequence, very much like~\citep{bahdanau15}. We discuss this
approach in details in Appendix~\ref{app:pointing}. In our experiments, we find
that utilizing one glimpse in the pointing mechanism yields performance gains
at an insignificant cost latency.

\section{\label{sec:training}Optimization with policy gradients}

\cite{vinyals2015pointer} proposes training a pointer network using a supervised
loss function comprising conditional log-likelihood, which factors into a cross
entropy objective between the network's output probabilities and the targets
provided by a TSP solver. Learning from examples in such a way is undesirable
for NP-hard problems because (1) the performance of the model is tied to the
quality of the supervised labels, (2) getting high-quality labeled data is
expensive and may be infeasible for new problem statements, (3) one cares more
about finding a competitive solution more than replicating the results of
another algorithm.

By contrast, we believe Reinforcement Learning (RL) provides an appropriate
paradigm for training neural networks for combinatorial optimization,
especially because these problems have relatively simple reward mechanisms that
could be even used at test time.  We hence propose to use model-free
policy-based Reinforcement Learning to optimize the parameters of a pointer
network denoted $\btheta$.  Our training objective is the expected tour length
which, given an input graph $s$, is defined as 
\begin{equation}
  J(\btheta \mid s) = \mathbb{E}_{\pi \sim p_{\theta}(.|s)} \,L(\pi \mid s)~.
  \label{eq:rl-obj}
\end{equation}
During training, our graphs are drawn from a distribution
$\mathcal{S}$, and the total training objective involves sampling from
the distribution of graphs, \ie~$J(\btheta) = \mathbb{E}_{s \sim \mathcal{S}} \,J(\btheta
\mid s)$~.

\begin{algorithm}[t]
  \centering
  \small
  \caption{\label{alg:actor_critic}Actor-critic training}
  \begin{algorithmic}[1]
    \Procedure{Train}{training set $S$, number of training steps $T$, batch size $B$}
      \State Initialize pointer network params $\theta$
      \State Initialize critic network params $\theta_v$
      \For{$t = 1$ to $T$}
        \State $s_i \sim \Call{SampleInput}{S}$ for $i \in \{1, \ldots, B\}$
        \State $\pi_i \sim \Call{SampleSolution}{p_{\theta}(.|s_i)}$ for $i \in \{1, \ldots, B\}$
        \State $b_i \gets b_{\theta_v}(s_i)$ for $i \in \{1, \ldots, B\}$
        \State $g_\theta \gets \frac{1}{B} \sum_{i=1}^{B} (L(\pi_i|s_i) - b_i) \nabla_\theta \log{p_\theta(\pi_i | s_i)}$
        \State $\mathcal{L}_{v} \gets \frac{1}{B} \sum_{i=1}^{B} \vectornorm{b_i - L(\pi_i)}_2^2$
        \State $\theta \gets \Call{Adam}{\theta, g_{\theta}}$
        \State $\theta_v \gets \Call{Adam}{\theta_v, \nabla_{\theta_v} \mathcal{L}_{v}}$
      \EndFor
  \State \Return $\theta$
  \EndProcedure
  \end{algorithmic}
\end{algorithm}

We resort to policy gradient methods and stochastic gradient descent
to optimize the parameters. The gradient of \eqref{eq:rl-obj} is
formulated using the well-known REINFORCE algorithm~\citep{reinforce}:
\begin{align}
  \label{eqn:rl_grad}
  \nabla_\theta J(\theta \mid s) = 
  \mathbb{E}_{\pi \sim p_{\theta}(.\mid s)}\Big[\big( L(\pi \mid s)-b(s) \big)\nabla_\theta 
    \log{p_{\theta}(\pi \mid s)}\Big]~,
\end{align}
where $b(s)$ denotes a baseline function that does not depend on $\pi$
and estimates the expected tour length to reduce the variance of the
gradients.

By drawing $B$ {\em i.i.d.} sample graphs $s_1, s_2, \dots, s_B \sim
\mathcal{S}$ and sampling a single tour per graph, \mbox{\ie~$\pi_i \sim
p_{\theta}(.\mid s_i)$}, the gradient in \eqref{eqn:rl_grad} is
approximated with Monte Carlo sampling as follows:
\begin{align}
  \label{eqn:mc}
  \nabla_\theta J(\theta)
  ~\approx~ \frac{1}{B} \sum_{i=1}^B 
  \Big(L(\pi_i|s_i)-b(s_i)\Big) \nabla_\theta \log{p_{\theta}(\pi_i \mid s_i)}~.
\end{align}

A simple and popular choice of the baseline $b(s)$ is an exponential
moving average of the rewards obtained by the network over time to
account for the fact that the policy improves with training. While
this choice of baseline proved sufficient to improve
upon the Christofides algorithm, it suffers from not being able 
to differentiate between different input graphs. 
In particular, the optimal tour $\pi^*$ for a difficult graph
$s$ may be still discouraged if $L(\pi^*|s) > b$ because $b$ is
shared across all instances in the batch.

Using a parametric baseline to estimate the expected
tour length $\mathbb{E}_{\pi \sim p_{\theta}(.|s)} L(\pi \mid s)$
typically improves learning. Therefore, we introduce an auxiliary
network, called a {\em critic} and parameterized by $\theta_v$,
to learn the expected tour length found by our current policy
$p_{\theta}$ given an input sequence $s$. 
The critic is trained with stochastic gradient descent on a mean squared
error objective between its predictions $b_{\theta_v}(s)$ and the
actual tour lengths sampled by the most recent policy. The additional
objective is formulated as 
\begin{align}
  \mathcal{L}(\theta_v) = \frac{1}{B} \sum_{i=1}^B
  \, \big\lVert \, b_{\theta_v}(s_i) - L(\pi_i \mid s_i) \, \big\rVert_2^2~.
\end{align}

\paragraph{Critic's architecture for TSP.} We now explain how our critic maps an input
sequence $s$ into a baseline prediction $b_{\theta_v}(s)$.
Our critic comprises three neural network modules: 1) an LSTM encoder, 2) an LSTM process
block and 3) a 2-layer ReLU neural network decoder.
Its encoder has the same architecture as that of our pointer network's encoder
and encodes an input sequence $s$ into a sequence of latent memory states 
and a hidden state $h$. The process block, similarly to ~\citep{vinyals2015order},
then performs P steps of computation over the hidden state $h$.
Each processing step updates this hidden state by glimpsing at the memory states 
as described in Appendix~\ref{app:pointing} and 
feeds the output of the glimpse function as input to the next processing step.
At the end of the process block, the obtained hidden state is then decoded into a 
baseline prediction (i.e a single scalar) by two fully connected layers 
with respectively d and 1 unit(s).

Our training algorithm, described in Algorithm~\ref{alg:actor_critic},
is closely related to the asynchronous advantage actor-critic (A3C)
proposed in \citep{a3c}, as the difference between the sampled tour
lengths and the critic's predictions is an unbiased estimate of the
advantage function.  We perform our updates asynchronously across
multiple workers, but each worker also handles a mini-batch of graphs
for better gradient estimates.

\subsection{\label{sec:search}Search Strategies}

As evaluating a tour length is inexpensive, our TSP agent can easily simulate a
search procedure at inference time by considering multiple candidate solutions
per graph and selecting the best.  This inference process resembles how solvers
search over a large set of feasible solutions.  In this paper, we consider two
search strategies detailed below, which we refer to as {\em sampling} and {\em
active search}.

\paragraph{Sampling.} Our first approach is simply to sample multiple candidate
tours from our stochastic policy $p_{\theta}(.|s)$ and select the shortest one.
In contrast to heuristic solvers, we do not enforce our model to sample
different tours during the process. However, we can control the diversity of the
sampled tours with a temperature hyperparameter when sampling from our
non-parametric softmax (see Appendix~\ref{app:temperature}). This sampling
process yields significant improvements over greedy decoding, which always
selects the index with the largest probability. We also considered perturbing
the pointing mechanism with random noise and greedily decoding from the
obtained modified policy, similarly to \citep{Cho2016napd}, but this proves less
effective than sampling in our experiments.

\paragraph{Active Search.}  Rather than sampling with a fixed model and
ignoring the reward information obtained from the sampled solutions, one can
refine the parameters of the stochastic policy $p_{\theta}$ during inference to
minimize $\mathbb{E}_{\pi \sim p_{\theta}(.\mid s)} L(\pi\mid s)$ on a {\em
single test input $s$}.  This approach proves especially competitive when
starting from a trained model. Remarkably, it also produces
satisfying solutions when starting from an untrained model. We refer to
these two approaches as {\em RL pretraining-Active Search} and {\em Active
Search} because the model actively updates its parameters while searching
for candidate solutions on a single test instance.

Active Search applies policy gradients similarly to
Algorithm~\ref{alg:actor_critic} but draws Monte Carlo samples over candidate
solutions $\pi_1 \dots \pi_B \sim p_{\theta}(\cdot | s)$ for a single test input. 
It resorts to an exponential moving average baseline, rather than a critic,
as there is no need to differentiate between inputs. Our Active Search training
algorithm is presented in Algorithm \ref{alg:active_search}.  We note that
while RL training does not require supervision, it still requires training data
and hence generalization depends on the training data distribution. In
contrast, Active Search is distribution independent.  Finally, since we encode
a set of cities as a sequence, we randomly shuffle the input sequence before
feeding it to our pointer network. This increases the stochasticity of the
sampling procedure and leads to large improvements in Active Search.

\begin{algorithm}[t]
  \centering
  \small
  \caption{\label{alg:active_search}Active Search}
  \begin{algorithmic}[1]
    \Procedure{ActiveSearch}{input s, $\theta$, number of candidates K, B, $\alpha$}
      \State $\pi \gets \Call{RandomSolution}$
      \State $L_{\pi} \gets L(\pi\mid s)$
      \State $n \gets \lceil \frac{K}{B} \rceil$
      \For{$t = 1 \dots n$}
        \State $\pi_i \sim \Call{SampleSolution}{p_{\theta}(.\mid s)}$ for $i \in \{1, \ldots, B\}$
        \State $j \gets \Call{Argmin}{L(\pi_1\mid s) \dots L(\pi_B\mid s)}$
        \State $L_j \gets L(\pi_j\mid s)$
        \If {$L_j < L_{\pi}$}
          \State $\pi \gets \pi_j$
          \State $L_{\pi} \gets L_j$
        \EndIf
        \State $g_\theta \gets \frac{1}{B} \sum_{i=1}^{B}
          (L(\pi_i\mid s) - b) \nabla_\theta \log{p_\theta(\pi_i \mid  s)}$
        \State $\theta \gets \Call{Adam}{\theta, g_{\theta}}$
        \State $b \gets \alpha \times b + (1-\alpha) \times (\frac{1}{B} \sum_{i=1}^B b_i)$
      \EndFor
  \State \Return $\pi$
  \EndProcedure
  \end{algorithmic}
\end{algorithm}

\section{\label{sec:experiments}Experiments}

We conduct experiments to investigate the behavior of the proposed Neural
Combinatorial Optimization methods. We consider three benchmark tasks,
Euclidean TSP20, 50 and 100, for which we generate a test set of $1,000$
graphs. Points are drawn uniformly at random in the unit square $[0,1]^2$.

\subsection{Experimental details}
Across all experiments, we use mini-batches of $128$ sequences, LSTM
cells with $128$ hidden units, and embed the two coordinates of each
point in a $128$-dimensional space. We train our models with the Adam
optimizer~\citep{adam} and use an initial learning rate of $10^{-3}$
for TSP20 and TSP50 and $10^{-4}$ for TSP100 that we decay every
$5000$ steps by a factor of $0.96$. We initialize our parameters
uniformly at random within $[-0.08, 0.08]$ and clip the $L2$ norm of
our gradients to $1.0$. We use up to one attention glimpse. When
searching, the mini-batches either consist of replications of the test
sequence or its permutations. The baseline decay is set to $\alpha =
0.99$ in Active Search. Our model and training code in
Tensorflow~\citep{tensorflow} will be made availabe soon.
\tabref{tab:configuration} summarizes the configurations and different
search strategies used in the experiments. The variations of our
method, experimental procedure and results are as follows.

\begin{table}[t]
    \centering
    \small
    \captionsetup{font=footnotesize}
    \caption{Different learning configurations.}
    \begin{tabular}{|l | c | c | c |}
        \hline
            \multirow{2}{*}{Configuration} & Learn on &  Sampling & Refining \\
                                           & training data  & on test set & on test set \\
        \hline
        RL pretraining-Greedy & Yes & No & No \\
        \hline
        Active Search (AS) &  No & Yes & Yes \\
        \hline
        RL pretraining-Sampling & Yes & Yes & No \\
        \hline
        RL pretraining-Active Search & Yes & Yes & Yes \\
        \hline
    \end{tabular}
    \label{tab:configuration}
\end{table}

\paragraph{Supervised Learning.~}
In addition to the described baselines, we implement and train a pointer
network with supervised learning, similarly to \citep{vinyals2015pointer}.
While our supervised data consists of one million optimal tours, we find that
our supervised learning results are not as good as those reported in
by~\citep{vinyals2015pointer}. We suspect that learning from optimal tours is
harder for supervised pointer networks due to subtle features that the model
cannot figure out only by looking at given supervised targets. We thus refer to
the results in \citep{vinyals2015pointer} for TSP20 and TSP50 and report our
results on TSP100, all of which are suboptimal compared to other approaches.

\paragraph{RL pretraining.~}
For the RL experiments, we generate training mini-batches of inputs on the fly
and update the model parameters with the Actor Critic Algorithm
\ref{alg:actor_critic}. We use a validation set of $10,000$ randomly generated
instances for hyper-parameters tuning. Our critic consists of an encoder
network which has the same architecture as that of the policy network, but
followed by $3$ processing steps and $2$ fully connected layers. We find
that clipping the logits to $[-10, 10]$ with a $\tanh(\cdot)$ activation
function, as described in Appendix~\ref{app:logit_clipping}, helps with
exploration and yields marginal performance gains. The simplest search strategy
using an RL pretrained model is greedy decoding, \ie~selecting the city with
the largest probability at each decoding step.
We also experiment with decoding greedily from a set of 16 pretrained models at inference time.
For each graph, the tour found by each individual model is collected and the shortest tour
is chosen. We refer to those approaches as {\em RL pretraining-greedy} 
and {\em RL pretraining-greedy@16}.

\paragraph{RL pretraining-Sampling.~}
For each test instance, we sample $1,280,000$ candidate solutions from a
pretrained model and keep track of the shortest tour. A grid search over the
temperature hyperparameter found respective temperatures of $2.0$, $2.2$ and
$1.5$ to yield the best results for TSP20, TSP50 and TSP100. We refer to the
tuned temperature hyperparameter as $T^*$. Since sampling does not require 
parameter udpates and is entirely parallelizable, we use a larger batch size 
for speed purposes.

\paragraph{RL pretraining-Active Search.~}
For each test instance, we initialize the model parameters from a pretrained RL
model and run Active Search for up to $10,000$ training steps with a batch
size of $128$, sampling a total of $1,280,000$ candidate solutions. 
We set the learning rate to a hundredth 
of the initial learning rate the TSP agent was
trained on (i.e. $10^{-5}$ for TSP20/TSP50 and $10^{-6}$ for TSP100).

\paragraph{Active Search.~}
We allow the model to train much longer to account for the fact that it starts
from scratch. For each test graph, we run Active Search for $100,000$ training
steps on TSP20/TSP50 and $200,000$ training steps on TSP100.

\subsection{\label{sec:tsp_results}Results and Analyses}
We compare our methods against 3 different baselines of increasing performance
and complexity: 
1) Christofides, 
2) the vehicle routing solver from OR-Tools~\citep{OR-Tools} and 
3) optimality.
Christofides solutions are obtained in polynomial time and guaranteed to be within
a $1.5$ ratio of optimality.
OR-Tools improves over Christofides' solutions with simple local search operators, 
including 2-opt~\citep{twoopt} and a version of the Lin-Kernighan heuristic~\citep{lk},
stopping when it reaches a local minimum. In order to escape poor local optima,
OR-Tools' local search can also be run in conjunction with different metaheuristics,
such as simulated annealing~\citep{simulated_annealing}, tabu search~\citep{tabubook} 
or guided local search~\citep{guidedlocalsearch}. 
OR-Tools' vehicle routing solver can tackle a superset of the TSP and operates 
at a higher level of generality than solvers that are highly specific to the TSP.
While not state-of-the art for the TSP, it is a common choice for general
routing problems and provides a reasonable baseline between the simplicity 
of the most basic local search operators and the sophistication of the strongest
solvers.
Optimal solutions are obtained via Concorde~\citep{concorde} and 
LK-H's local search ~\citep{lkhsolver, lkh}. While only Concorde provably solves 
instances to optimality, we empirically find that LK-H also achieves optimal 
solutions on all of our test sets.

\begin{table}[t]
    \centering
    \small
    \captionsetup{font=footnotesize}
    \caption{Average tour lengths (lower is better). Results marked
    ${}^{(\dagger)}$ are from \citep{vinyals2015pointer}.}
    \label{tab:results}
    \begin{tabular}{|l || c | c | c | c | c | c || c | c | c |}
        \hline
        \multirow{2}{*}{Task} & \multirow{2}{0.1\linewidth}{\centering Supervised Learning} &
        \multicolumn{4}{c|}{RL pretraining} & \multirow{2}{*}{AS} & Christo &
        OR Tools' & \multirow{2}{*}{Optimal} \\
        \cline{3-6}
        & & greedy & greedy@16 & sampling & AS & & -fides & local search & \\
        \hline
        TSP20  & $3.88^{(\dagger)}$ & 3.89 & $-$  & 3.82 & 3.82 & 3.96 & 4.30 & 3.85 & 3.82 \\
        TSP50  & $6.09^{(\dagger)}$ & 5.95 & 5.80 & 5.70 & 5.70 & 5.87 & 6.62 & 5.80 & 5.68 \\
        TSP100 & 10.81              & 8.30 & 7.97 & 7.88 & 7.83 & 8.19 & 9.18 & 7.99 & 7.77 \\
        \hline
    \end{tabular}
\end{table}

\begin{table}[t]
  \centering
  \small
  \captionsetup{font=footnotesize}
  \caption{\label{tab:runtime}Running times in seconds (s) of greedy methods
  compared to OR Tool's local search and solvers that find the optimal solutions. 
  Time is measured over the entire test set and averaged.}
  \begin{tabular}{| l || c | c || c | c | c |}
    \hline
        \multirow{2}{*}{Task}  
      & \multicolumn{2}{c|}{RL pretraining} 
      & OR-Tools'
      & \multicolumn{2}{c|}{Optimal} \\
    \cline{2-3} \cline{5-6}
      & greedy & greedy@16 & local search & Concorde & LK-H \\
    \hline \hline
    TSP50  & $0.003$s & $0.04$s & $0.02s$ & $0.05$s & $0.14$s \\
    TSP100 & $0.01$s  & $0.15$s & $0.10s$ & $0.22$s & $0.88$s \\
    \hline
  \end{tabular}
\end{table}

We report the average tour lengths of our approaches on TSP20, TSP50, and
TSP100 in~\tabref{tab:results}. 
Notably, results demonstrate that training with
RL significantly improves over supervised learning \citep{vinyals2015pointer}.
All our methods comfortably surpass Christofides' heuristic, 
including RL pretraining-Greedy which also does not rely on search. 
Table~\ref{tab:runtime} compares the running times of our greedy methods 
to the aforementioned baselines, with our methods running on a single Nvidia 
Tesla K80 GPU, Concorde and LK-H running on an Intel Xeon CPU E5-1650 v3 3.50GHz CPU
and OR-Tool on an Intel Haswell CPU. We find that both greedy approaches are
time-efficient and just a few percents worse than optimality.

Searching at inference time proves crucial to get closer to optimality but comes
at the expense of longer running times. Fortunately, the search from RL pretraining-Sampling 
and RL pretraining-Active Search can be stopped early with a small performance
tradeoff in terms of the final objective. This can be seen in Table~\ref{tab:perf},
where we show their performances and corresponding running times
as a function of how many solutions they consider.

We also find that many of our RL pretraining methods outperform OR-Tools' local search,
including RL pretraining-Greedy@16 which runs similarly fast.
Table~\ref{tab:meta} in Appendix~\ref{sec:tsp_baselines} 
presents the performance of the metaheuristics
as they consider more solutions and the corresponding running times.
In our experiments, Neural Combinatorial proves superior than Simulated Annealing
but is slightly less competitive that Tabu Search and much less so than Guided Local Search.

\begin{figure}[t]
    \centering
    \small
    \includegraphics[width=0.48\textwidth]{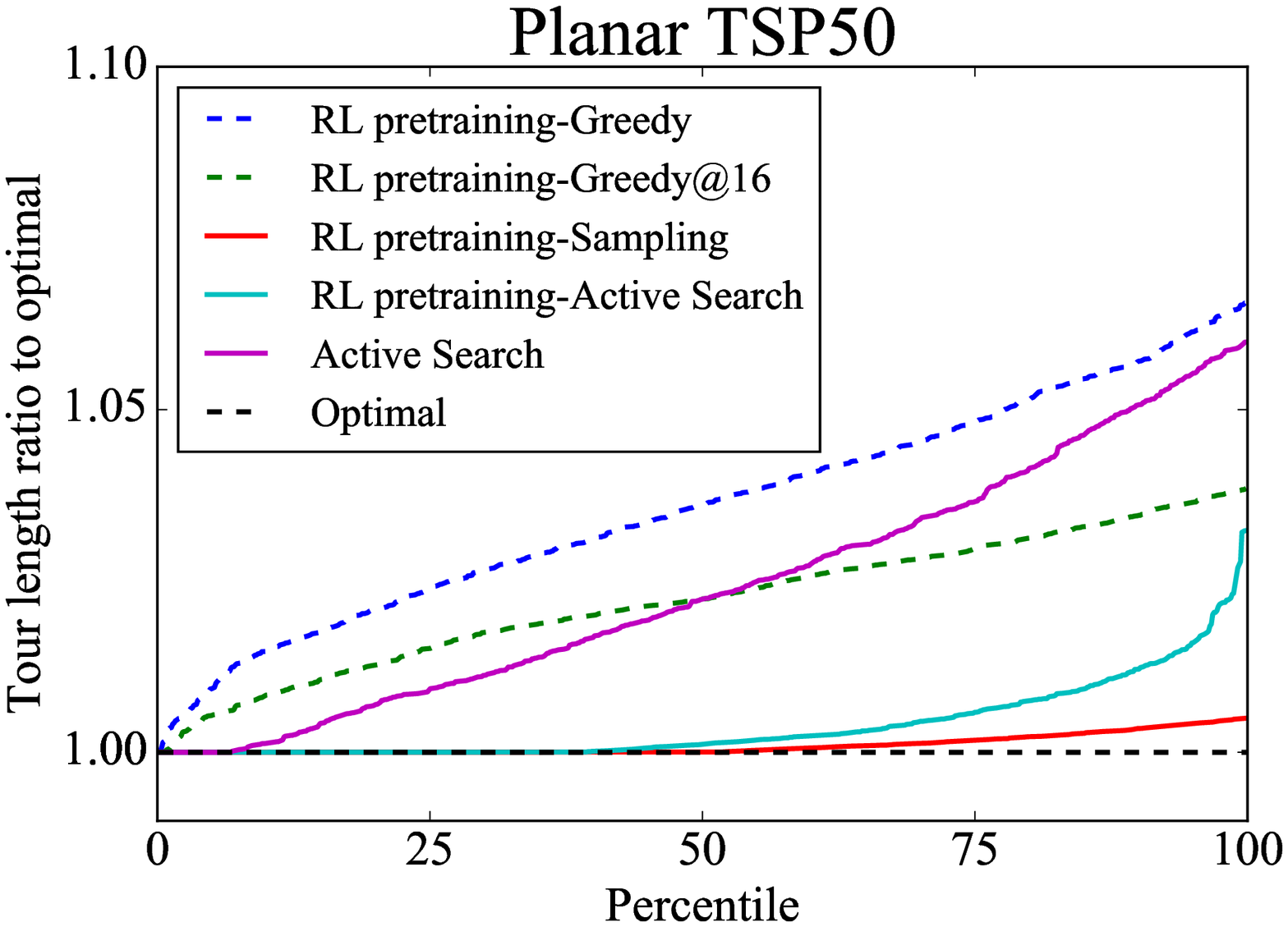}
    \includegraphics[width=0.48\textwidth]{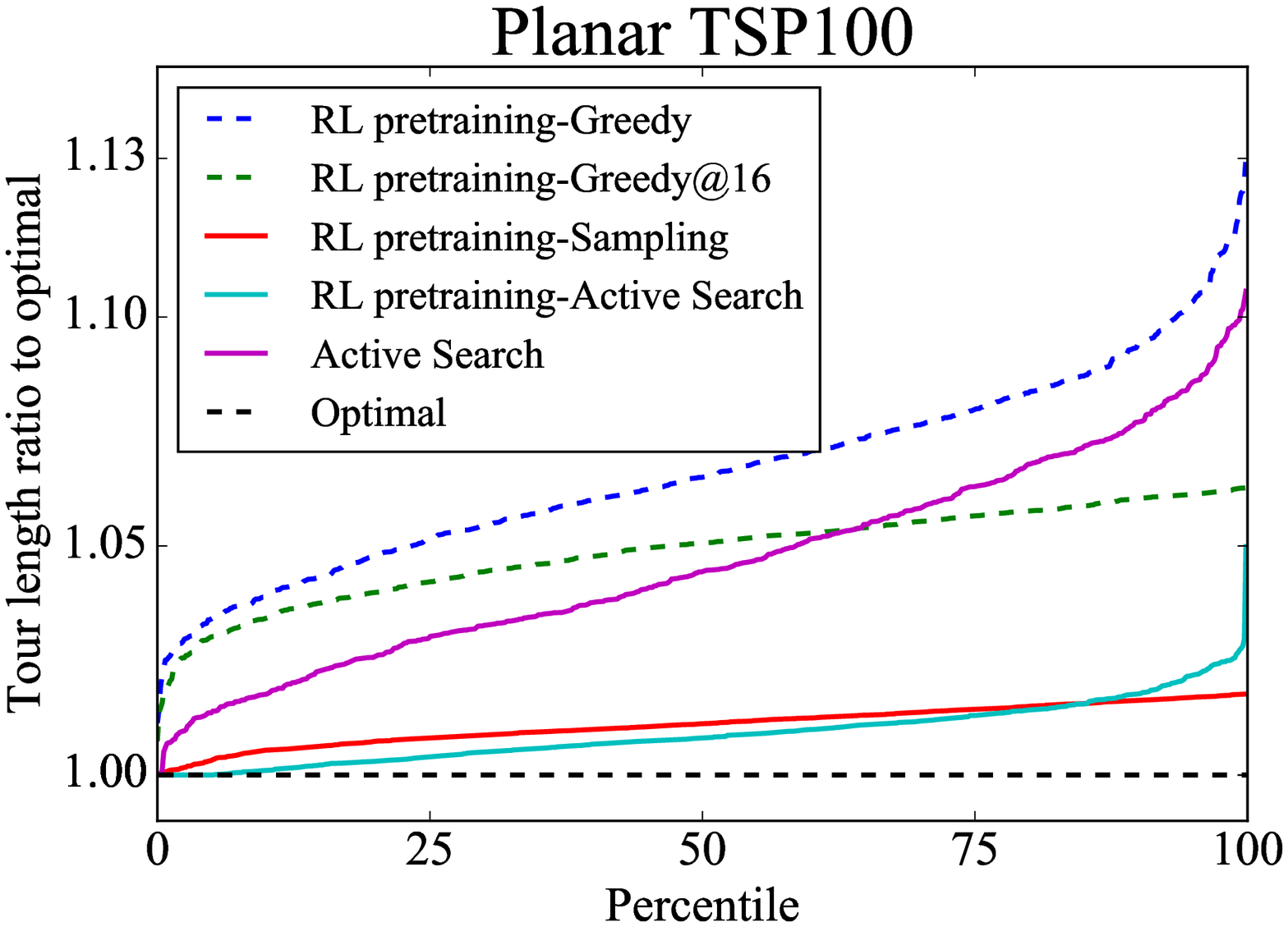}
    \caption{Sorted tour length ratios to optimality}
    \label{fig:tour_diffs}
\end{figure}

We present a more detailed comparison of our methods in
Figure~\ref{fig:tour_diffs}, where we sort the ratios to optimality of our
different learning configurations. 
RL pretraining-Sampling and RL pretraining-Active Search are the most competitive
Neural Combinatorial Optimization methods and recover the optimal solution 
in a significant number of our test cases. 
We find that for small solution spaces, RL pretraining-Sampling, 
with a finetuned softmax temperature, outperforms RL pretraining-Active Search
with the latter sometimes orienting the search towards suboptimal 
regions of the solution space
(see TSP50 results in Table~\ref{tab:perf} and Figure~\ref{fig:tour_diffs}).
Furthermore, RL pretraining-Sampling benefits from being fully parallelizable 
and runs faster than RL pretraining-Active Search.
However, for larger solution spaces, RL-pretraining Active Search 
proves superior both when controlling for the number of sampled solutions 
or the running time.
Interestingly, Active Search - which starts from an untrained model - 
also produces competitive tours but requires a considerable amount of time 
(respectively 7 and 25 hours per instance of TSP50/TSP100).
Finally, we show randomly picked example tours found by our methods in
Figure~\ref{fig:tours} in Appendix~\ref{sec:sample_tours}.make

\begin{table}[t]
  \centering
  \small
  \captionsetup{font=footnotesize}
  \caption{Average tour lengths of RL pretraining-Sampling 
  and RL pretraining-Active Search as they sample more solutions. 
  Corresponding running times on a single Tesla K80 GPU are in parantheses.}
  \begin{tabular}{| l || l | c | c | c |}
    \hline
    \multirow{2}{*}{Task} & \multirow{2}{*}{\# Solutions} & \multicolumn{3}{c|}{RL pretraining} \\
    \cline{3-5}
    & & Sampling $T=1$ & Sampling $T=T^*$ & Active Search \\
    \hline \hline
    \multirow{6}{*}{TSP50}  & 128        & 5.80 (3.4s)   & 5.80 (3.4s)   & 5.80 (0.5s) \\
                            & 1,280      & 5.77 (3.4s)   & 5.75 (3.4s)   & 5.76 (5s) \\
                            & 12,800     & 5.75 (13.8s)  & 5.73 (13.8s)  & 5.74 (50s) \\
                            & 128,000    & 5.73 (110s)   & 5.71 (110s)   & 5.72 (500s) \\
                            & 1,280,000  & 5.72 (1080s)  & 5.70 (1080s)  & 5.70 (5000s) \\
    \hline \hline
    \multirow{6}{*}{TSP100} & 128        & 8.05 (10.3s)  & 8.09 (10.3s)  & 8.04 (1.2s) \\
                            & 1,280      & 8.00 (10.3s)  & 8.00 (10.3s)  & 7.98 (12s) \\
                            & 12,800     & 7.95 (31s)    & 7.95 (31s)    & 7.92 (120s) \\
                            & 128,000    & 7.92 (265s)   & 7.91 (265s)   & 7.87 (1200s) \\
                            & 1,280,000  & 7.89 (2640s)  & 7.88 (2640s)  & 7.83 (12000s) \\
    \hline
  \end{tabular}
  \label{tab:perf}
\end{table}

\section{\label{sec:extensions}Generalization to other problems}

In this section, we discuss how to apply Neural Combinatorial Optimization to
other problems than the TSP. 
In Neural Combinatorial Optimization, the model architecture 
is tied to the given combinatorial optimization problem. Examples
of useful networks include the pointer network, when the output is a
permutation or a truncated permutation or a subset of the input, and the
classical seq2seq model for other kinds of structured outputs. For
combinatorial problems that require to assign labels to elements of the input,
such as graph coloring, it is also possible to combine a pointer module and a
softmax module to simultaneously point and assign at decoding time.
Given a model that encodes an instance of a given combinatorial optimization task
and repeatedly branches into subtrees to construct a solution, 
the training procedures described in Section~\ref{sec:training} can then be applied 
by adapting the reward function depending on the optimization problem being considered.

Additionally, one also needs to ensure the feasibility of the obtained solutions.
For certain combinatorial problems, it is straightforward to know exactly which
branches do not lead to any feasible solutions at decoding time. 
We can then simply manually assign them a zero probability when decoding, 
similarly to how we enforce our model to not point at the same city 
twice in our pointing mechanism (see Appendix~\ref{app:pointing}).
However, for many combinatorial problems, coming up with a feasible solution
can be a challenge in itself. Consider, for example, the Travelling Salesman Problem
with Time Windows, where the travelling salesman has the additional constraint
of visiting each city during a specific time window. It might be that most branches
being considered early in the tour do not lead to any solution that respects all time windows.
In such cases, knowing exactly which branches are feasible requires searching 
their subtrees, a time-consuming process that is not much easier than directly
searching for the optimal solution unless using problem-specific heuristics.

Rather than explicitly constraining the model to only sample feasible solutions, 
one can also let the model learn to respect the problem's constraints. A simple approach,
to be verified experimentally in future work, consists in augmenting the 
objective function with a term that penalizes solutions for violating 
the problem's constraints, similarly to penalty methods in constrained optimization. 
While this does not guarantee that the model consistently samples feasible solutions
at inference time, this is not necessarily problematic as we can simply ignore 
infeasible solutions and resample from the model (for RL pretraining-Sampling
and RL-pretraining Active Search).
It is also conceivable to combine both approaches by assigning zero probabilities
to branches that are easily identifiable as infeasible while still penalizing
infeasible solutions once they are entirely constructed.

\subsection{KnapSack example}
As an example of the flexibility of Neural Combinatorial Optimization, we
consider the KnapSack problem, another intensively studied problem in computer
science.  Given a set of $n$ items $i = 1 ... n$, each with weight $w_i$ and
value $v_i$ and a maximum weight capacity of $W$, the 0-1 KnapSack problem
consists in maximizing the sum of the values of items present in the knapsack
so that the sum of the weights is less than or equal to the knapsack capacity:
\begin{align}
  \begin{aligned}
    & \underset{S \subseteq \{1, 2, ..., n\}}{\text{max}} & & \sum_{i \in S} v_i \\
    & \text{subject to} & & \sum_{i \in S} w_i \leq W
  \end{aligned}
\end{align}
With $w_i$, $v_i$ and $W$ taking real values, the problem is
NP-hard~\citep{knapsack}. A simple yet strong heuristic is to take the items
ordered by their weight-to-value ratios until they fill up the weight capacity.
We apply the pointer network and encode each knapsack instance as a sequence of
2D vectors $(w_i, v_i)$. At decoding time, the pointer network points to items
to include in the knapsack and stops when the total weight of the items
collected so far exceeds the weight capacity.

We generate three datasets, KNAP50, KNAP100 and KNAP200, of a thousand
instances with items' weights and values drawn uniformly at random in $[0, 1]$.
Without loss of generality (since we can scale the items' weights), we set the
capacities to $12.5$ for KNAP50 and $25$ for KNAP100 and KNAP200.  We present
the performances of RL pretraining-Greedy and Active Search (which we run for
$5,000$ training steps) in Table~\ref{tab:knapsack} and compare them to two
simple baselines: the first baseline is the greedy weight-to-value ratio
heuristic; the second baseline is random search, where we sample as many
feasible solutions as seen by Active Search.  RL pretraining-Greedy yields
solutions that, in average, are just $1\%$ less than optimal and Active Search
solves all instances to optimality. 

\begin{table}[H]
  \centering
  \small
  \captionsetup{font=footnotesize}
  \caption{\small Results of RL pretraining-Greedy and Active Search on KnapSack (higher is better).}
  \begin{tabular}{| l || c | c || c | c | c |}
    \hline
    \multirow{2}{*}{Task} & RL pretraining & \multirow{2}{*}{Active Search} & \multirow{2}{*}{Random Search} & \multirow{2}{*}{Greedy} & \multirow{2}{*}{Optimal} \\
     & greedy &  &  &  & \\
    \hline \hline
    KNAP50   & 19.86 & \textbf{20.07} & 17.91 & 19.24 & \textbf{20.07} \\
    KNAP100  & 40.27 & \textbf{40.50} & 33.23 & 38.53 & \textbf{40.50} \\
    KNAP200  & 57.10 & \textbf{57.45} & 35.95 & 55.42 & \textbf{57.45} \\
    \hline
  \end{tabular}
  \label{tab:knapsack}
\end{table}

\section{\label{sec:conclusion}Conclusion}

This paper presents Neural Combinatorial Optimization, a framework to tackle
combinatorial optimization with reinforcement learning and neural networks.  We
focus on the traveling salesman problem (TSP) and present a set of results for
each variation of the framework.  Experiments demonstrate that Neural
Combinatorial Optimization achieves close to optimal results on 2D Euclidean
graphs with up to $100$ nodes.

\section*{Acknowledgments}
The authors would like to thank Vincent Furnon, Oriol Vinyals, Barret Zoph,
Lukasz Kaiser, Mustafa Ispir and the Google Brain team for insightful comments
and discussion.

\bibliography{iclr2017_conference}
\bibliographystyle{iclr2017_conference}

\newpage
\appendix
\section{\label{appendix} Appendix}

\subsection{Pointing and Attending} \label{app:pointing}
\paragraph{Pointing mechanism:}
Its computations are parameterized by two attention matrices 
$W_{ref}, W_{q} \in \mathbb{R}^{d \times d}$ 
and an attention vector $v \in \mathbb{R}^{d}$ as follows:
\begin{align}
  \label{eqn:attn_logits}
  &u_i = \begin{cases}
          v^{\top} \cdot \tanh{\left( W_{ref} \cdot r_i + W_{q} \cdot q \right)} 
          &\text{if $i \neq \pi(j)$ for all $j < i$} \\
          -\infty &\text{otherwise}
        \end{cases}
  \text{for $i = 1, 2, ..., k$} \\
  \label{eqn:attn_logits2}
  &A(ref, q; W_{ref}, W_q, v) \defeq softmax(u).
\end{align}

Our pointer network, at decoder step $j$, then assigns the probability of
visiting the next point $\pi(j)$ of the tour as follows:
\begin{align}
  p(\pi(j) | \pi(<j), s) \defeq A(enc_{1:n}, dec_j).
\end{align}

Setting the logits of cities that already appeared in the tour to $-\infty$, as
shown in Equation \ref{eqn:attn_logits}, ensures that our model only points at
cities that have yet to be visited and hence outputs valid TSP tours.

\paragraph{Attending mechanism:}
Specifically, our glimpse function $G(ref, q)$ takes the same inputs as the
attention function $A$ and is parameterized by $W^g_{ref}, W^{g}_q \in
\mathbb{R}^{d \times d}$ and $v^{g} \in \mathbb{R}^d$. It performs the
following computations:
\begin{align}
  &p = A(ref, q; W^{g}_{ref}, W^{g}_{q}, v^{g}) \\
  &G(ref, q; W^{g}_{ref}, W^{g}_{q}, v^{g}) \defeq \sum_{i=1}^{k} r_i p_i.
\end{align}

The glimpse function $G$ essentially computes a linear combination of the
reference vectors weighted by the attention probabilities. It can also be
applied multiple times on the same reference set $ref$:
\begin{align}
  &g_0 \defeq q \\
  &g_l \defeq G(ref, g_{l-1}; W^{g}_{ref}, W^{g}_{q}, v^{g})
\end{align}

Finally, the ultimate $g_l$ vector is passed to the attention function $A(ref,
g_l; W_{ref}, W_{q}, v)$ to produce the probabilities of the pointing
mechanism. We observed empirically that glimpsing more than once with the same
parameters made the model less likely to learn and barely improved the results.

\subsection{Improving exploration}
\paragraph{Softmax temperature: }
\label{app:temperature}
We modify Equation \ref{eqn:attn_logits2} as follows: 
\begin{equation}
A(ref, q, T; W_{ref}, W_q, v) \defeq softmax(u / T),
\end{equation}
where $T$ is a \textit{temperature} hyperparameter set to $T = 1$ during
training. When $T > 1$, the distribution represented by $A(ref, q)$ becomes
less steep, hence preventing the model from being overconfident.

\paragraph{Logit clipping: }
\label{app:logit_clipping}
We modify Equation \ref{eqn:attn_logits2} as follows: 
\begin{equation}
A(ref, q; W_{ref}, W_q, v) \defeq softmax(C \tanh(u)),
\end{equation}
where $C$ is a hyperparameter that controls the range of the logits and hence
the entropy of $A(ref, q)$.

\subsection{\label{sec:tsp_baselines}OR Tool's Metaheuristics Baselines for TSP}
\begin{table}[H]
  \centering
  \small
  \captionsetup{font=footnotesize}
  \caption{Performance of OR-Tools' metaheuristics as they consider more solutions.
  Corresponding running times in seconds (s) on a single Intel Haswell CPU 
  are in parantheses.}
  \begin{tabular}{| l | l | c | c | c |}
    \hline
    Task & \#Solutions & Simulated Annealing & Tabu Search & Guided Local Search \\
    \hline \hline
    \multirow{6}{*}{TSP50}  & 1          & 6.62 (0.03s)   & 6.62 (0.03s)   & 6.62 (0.03s) \\
                            & 128        & 5.81 (0.24s)   & 5.79 (3.4s)   & 5.76 (0.5s) \\
                            & 1,280      & 5.81 (4.2s)    & 5.73 (36s)   & 5.69 (5s) \\
                            & 12,800     & 5.81 (44s)     & 5.69 (330s)  & 5.68 (48s) \\
                            & 128,000    & 5.81 (460s)    & 5.68 (3200s) & 5.68 (450s) \\
                            & 1,280,000  & 5.81 (3960s)   & 5.68 (29650s)  & 5.68 (4530s) \\
    \hline \hline
    \multirow{6}{*}{TSP100} & 1          & 9.18 (0.07s)   & 9.18 (0.07s)  & 9.18 (0.07s) \\
                            & 128        & 8.00 (0.67s)   & 7.99 (15.3s)  & 7.94 (1.44s) \\
                            & 1,280      & 7.99 (15.7s)   & 7.93 (255s)  & 7.84 (18.4s) \\
                            & 12,800     & 7.99 (166s)    & 7.84 (2460s)  & 7.77 (182s) \\
                            & 128,000    & 7.99 (1650s)   & 7.79 (22740s)  & 7.77 (1740s) \\
                            & 1,280,000  & 7.99 (15810s)  & 7.78 (208230s)  & 7.77 (16150s) \\
    \hline
  \end{tabular}
  \label{tab:meta}
\end{table}

\subsection{\label{sec:sample_tours} Sample tours}
\begin{figure}[H]
    \centering
    \includegraphics[width=0.9\textwidth]{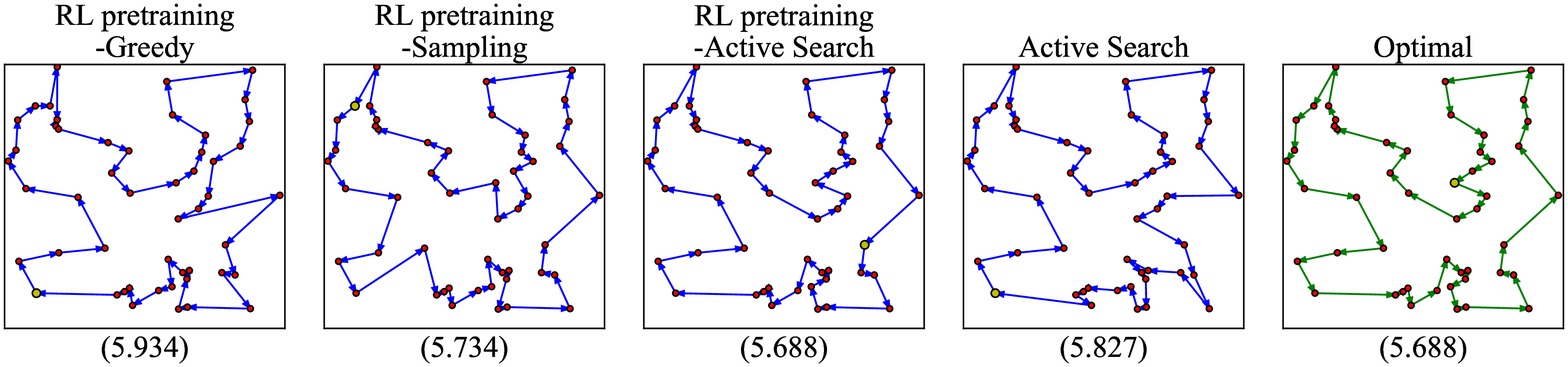}
    \includegraphics[width=0.9\textwidth]{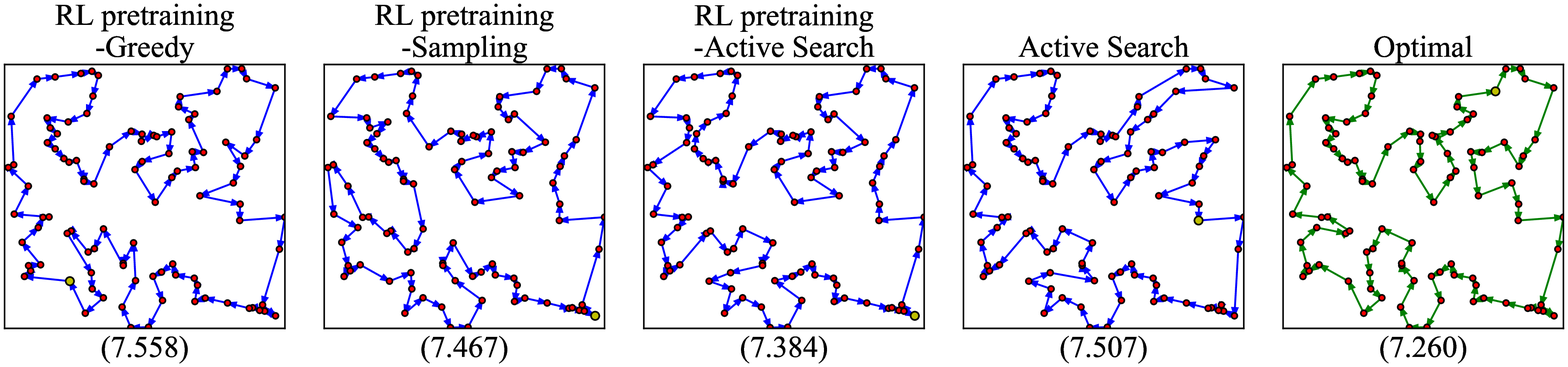}
    \caption{Sample tours. Top: TSP50; Bottom: TSP100.}
    \label{fig:tours}
\end{figure}

\end{document}